%% file: main.tex
\PassOptionsToPackage{type1,nofontspec}{XCharter}
\documentclass[11pt, a4paper, copyright, google]{google}

\usepackage[authoryear, sort&compress, round]{natbib}
\usepackage{wrapfig}
\usepackage{algorithm}    
\usepackage{algpseudocode}  
\usepackage[section]{placeins}
\usepackage{iftex}

\ifPDFTeX\else
  \microtypesetup{tracking=false}
\fi

\newcolumntype{Y}{>{\centering\arraybackslash}X}
\definecolor{shiftrow}{RGB}{232,242,252}

\uselogo{} 

\title{Dynamic Harness Search: Building Multi-Agent Systems Per-Query via Prediction}

\correspondingauthor{ssagar6@asu.edu, \{shashali, hejiec, soarik\}@google.com}

\reportnumber{0001} % Leave blank if n/a

\author[1,2*]{Som Sagar}
\author[1]{Shasha Li}
\author[1]{Hejie Cui}
\author[2]{Ransalu Senanayake}
\author[1]{Sercan \"{O}. Ar{\i}k}

\affil[1]{\thepa{}{}}
\affil[2]{Arizona State University}

\begin{abstract}
Agent harnesses specify the roles, instructions, tools, and communication structure used to solve a task, and the right harness depends on the query. Because the value of each design choice is observable only through execution, tailoring a harness to each query has required either executing alternatives at inference time or costly manual design. We introduce SHIFT, which moves execution out of the per-query search loop. A local LLM architect learns a policy over harness-building actions from search, and a value function that predicts, from measured executions, a utility balancing accuracy against execution cost. For each query, Monte Carlo tree search uses these predictions to construct a harness. Across 9,193 tasks in six benchmarks, from math to document and general-assistant tasks, with a Gemini 3.5 Flash executor, SHIFT attains the highest mean accuracy, about 80\%, outperforming 17 baselines that span prompting, prompt optimization, and workflow search, and exceeding the strongest baseline by 7.2 percentage points. A cheaper mode of SHIFT also attains a higher mean accuracy than every baseline while using 32\% fewer execution tokens than the strongest baseline. We further show that choosing structure, instructions, and tools jointly beats choosing only instructions or only tools by up to 9.1 percentage points, and that learned value selection identifies more accurate harnesses with lower execution cost from candidate pools.
\end{abstract}

\begin{document}

\maketitle

\input{sections/introduction}

\input{sections/related}

\input{sections/method}

\input{sections/experiment}

\input{sections/conclusion}

\FloatBarrier
\bibliography{main}

\input{sections/appendix}

\end{document}

%% file: sections/introduction.tex
\section{Introduction}

Consider an assistant that handles requests over an organization's documents. One user asks for a number stated in a supplied report; another asks the assistant to reconcile a budget workbook and return a corrected file. The first request may need only a lookup. The second requires reading several sheets, performing calculations, editing cells, and checking the result, work that can be divided among a planner, a solver, and a verifier~\cite{yao2022react,wu2023autogen}. Building such a system means deciding which agents to use, what instructions each follows, which tools it may call, and how agents pass results to one another. We call this specification the \emph{harness}. Yet the planner and verifier that help with the workbook may add unnecessary model calls and latency to the lookup. The right harness therefore depends on the query: it must provide the capabilities the task requires while keeping execution cost low.

Choosing this balance is difficult because harness components interact. A verifier's usefulness depends on the solver's output and the instructions for checking it; access to a tool helps only if the agent uses it effectively. These interactions make a component's value depend on both the query and the rest of the harness. Evaluating alternative designs through execution, however, adds to the cost of answering each query. Existing methods address this expense in different ways: workflow optimization searches offline for a reusable structure~\cite{zhang2025aflow,zhuge2024language}, while prompt optimization tunes instructions within a fixed program~\cite{khattab2023dspy,opsahl2024mipro}. Query-adaptive controllers generate workflows directly~\cite{zhang2025maas,taparia2026learning}, routers select among predefined models or workflows~\cite{chen2023frugalgpt,ong2025routellm,yue2025masrouter}, and learned quality predictors guide offline workflow search~\cite{shang2024agentsquare}. These approaches motivate our question: can a learned estimate of harness utility guide search over interacting design choices for each query while balancing accuracy against execution cost, allowing alternatives to be compared before any is executed?

\begin{figure}[tbp]
  \centering
  \includegraphics[width=\textwidth]{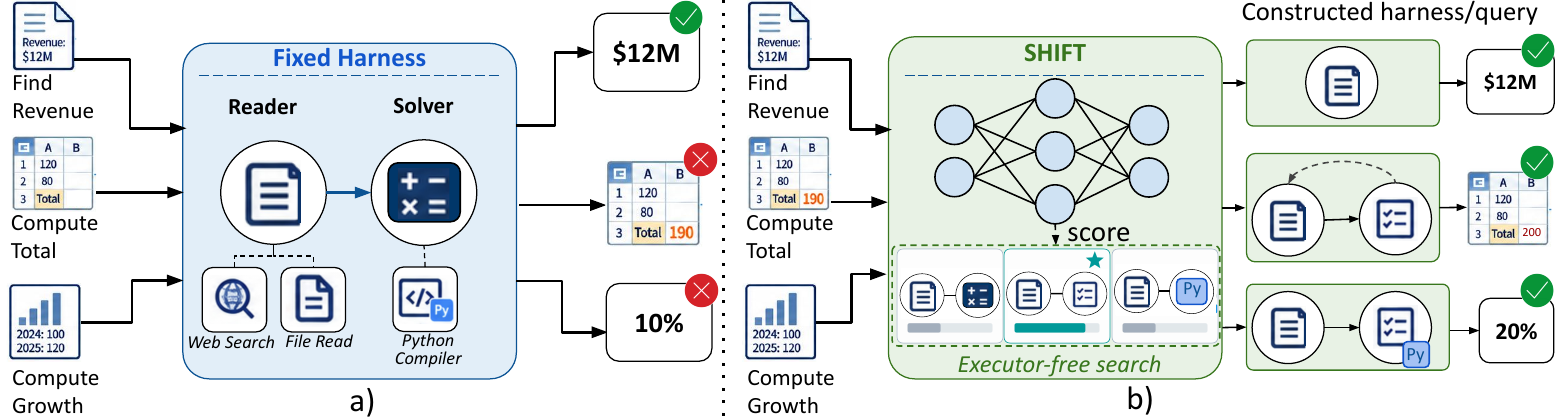}
\caption{\textbf{Tailoring the harness to the query.} Three queries ask for a reported revenue, a column total, and a growth rate. (a) A fixed harness runs the same Reader and Solver, with all their tools, on every query: it spends this structure on a simple lookup and still gets both calculations wrong (190 instead of 200, 10\% instead of 20\%). (b) SHIFT constructs a harness for each query, for example a single reader for the lookup, a reader with a verifier loop for the total, and a reader and a verifier with Python for the growth rate. The architect scores candidates without executing them, and only the constructed harness is run; in this example, all three answers are correct.}

  \label{fig:intro}
\end{figure}

We introduce SHIFT (Searching Harnesses In Forward-pass Trees), a framework that constructs an agent harness for each query without executing candidate designs during search. At its core is a small local model, the \emph{architect}, which proposes harness-building actions and predicts a utility balancing accuracy against execution cost. Monte Carlo tree search uses these predictions to explore combinations of agents, instructions, and tool permissions through architect inference alone (Figure~\ref{fig:intro}). The architect learns from search decisions and measured execution outcomes during training, whose cost is amortized over subsequent queries. At inference time, only the selected harness is executed.

We evaluate SHIFT on six benchmarks spanning math, code, multi-hop and document question answering, spreadsheets, and general assistance, against 17 baselines that share the same executor (Table~\ref{tab:main}). SHIFT-search selects the harness preferred by search and achieves the highest mean accuracy, 79.9\%, exceeding the strongest baseline by 7.2 percentage points. Its largest gains occur on OfficeQA and GAIA, where it improves accuracy over that baseline by 30.0 and 15.0 percentage points, respectively, while remaining within 1.2 percentage points of the best-performing methods on the other four benchmarks. We also introduce a lower-cost variant, SHIFT-value which selects using predicted utility and achieves 74.5\% mean accuracy, exceeding every baseline while using 32\% fewer execution tokens than the strongest baseline.

Our contributions are as follows:

\begin{enumerate}[noitemsep, topsep=0pt]
\item We formulate per-query harness construction as tree search over agent, instruction, and tool actions, guided by a learned architect that predicts the harness utility without executing it.
\item On six benchmarks with a shared executor, SHIFT-search is more accurate than all 17 baselines, and SHIFT-value exceeds them all in mean accuracy at 32\% lower token cost.
\item We show that optimizing structure, instructions, and tools jointly beats any one alone, and the learned architect picks accurate, low-cost harnesses and transfers to harder tasks.
\end{enumerate}

%% file: sections/related.tex
\section{Related Work}
\label{sec:related}

Prior work improves agent systems through reasoning and prompt design, automated workflow optimization, and query-adaptive composition. We review these directions with particular attention to how candidate designs are evaluated and when adaptation occurs.

\textbf{Reasoning and prompt optimization.} ReAct interleaves reasoning with actions so that agents can respond to tool observations~\cite{yao2022react}. Self-Refine uses iterative feedback to revise outputs~\cite{madaan2023selfrefine}, while Tree of Thoughts and Graph of Thoughts organize intermediate reasoning into branches or graphs that support exploring and comparing possible solutions~\cite{yao2023tree,besta2024graph}. Multi-agent frameworks distribute tasks among agents through role assignments and communication protocols~\cite{wu2023autogen,hong2024metagpt,chen2024agentverse}. The effectiveness of these reasoning and coordination mechanisms also depends on their instructions. Prompt optimization improves the instructions and few-shot examples of a fixed program using evaluation feedback, as in DSPy, MIPRO, and GEPA~\cite{khattab2023dspy,opsahl2024mipro,agrawal2025gepa}. SHIFT instead makes the harness itself the object of search before execution, choosing its agent structure, instructions, and tool permissions jointly as interacting design choices.

\textbf{Workflow optimization and surrogate evaluation.} Automating workflow design requires a way to judge whether a proposed configuration is worth keeping. Execution feedback guides revisions to code, agent graphs, and instructions~\cite{zhang2025aflow,zhuge2024language,cheng2024trace,hu2024adas}; DistIL uses feedback to distill a teacher into a response policy~\cite{agrawal2026distil}. When optimization produces a reusable design, its evaluation cost can be spread across subsequent queries. Constructing a new harness for each query makes repeated candidate execution more costly, motivating learned estimates of candidate quality. These estimates serve as surrogates for the measured outcomes of running a candidate. AgentSquare uses an LLM performance predictor to assess module recombinations~\cite{shang2024agentsquare}; however, newly evolved modules are still evaluated by execution in the task environment. SHIFT trains a local utility predictor on executed harnesses and uses its predictions to guide a new tree search for each query.

\textbf{Query-adaptive composition and routing.} The best agent configuration can vary with the query, motivating controllers that allocate computation or compose workflows per query. Model routers choose among language models~\cite{chen2023frugalgpt,ong2025routellm}, while controllers for multi-agent systems also select collaboration modes, roles, and model assignments~\cite{yue2025masrouter}. Adaptation can extend to the workflow itself: MaAS samples query-conditioned operator compositions from a supernet~\cite{zhang2025maas}. Other approaches generate communication topologies or executable multi-agent systems~\cite{li2026argdesigner,ye2025masgpt}, or configure workflows together with tools, token budgets, and prompts~\cite{taparia2026learning}. SHIFT combines query-specific construction with explicit lookahead over agent, instruction, and tool choices. Lookahead considers how an action affects the harness built by subsequent actions. A learned policy proposes actions, while a utility predictor trained on execution outcomes evaluates candidate harnesses during Monte Carlo tree search. This allows SHIFT to compare alternative designs for each query without executing candidates during inference-time search.

%% file: sections/method.tex
\section{Methodology}
\label{sec:method}

\begin{figure}[tbp]
    \centering
    \includegraphics[width=1\linewidth]{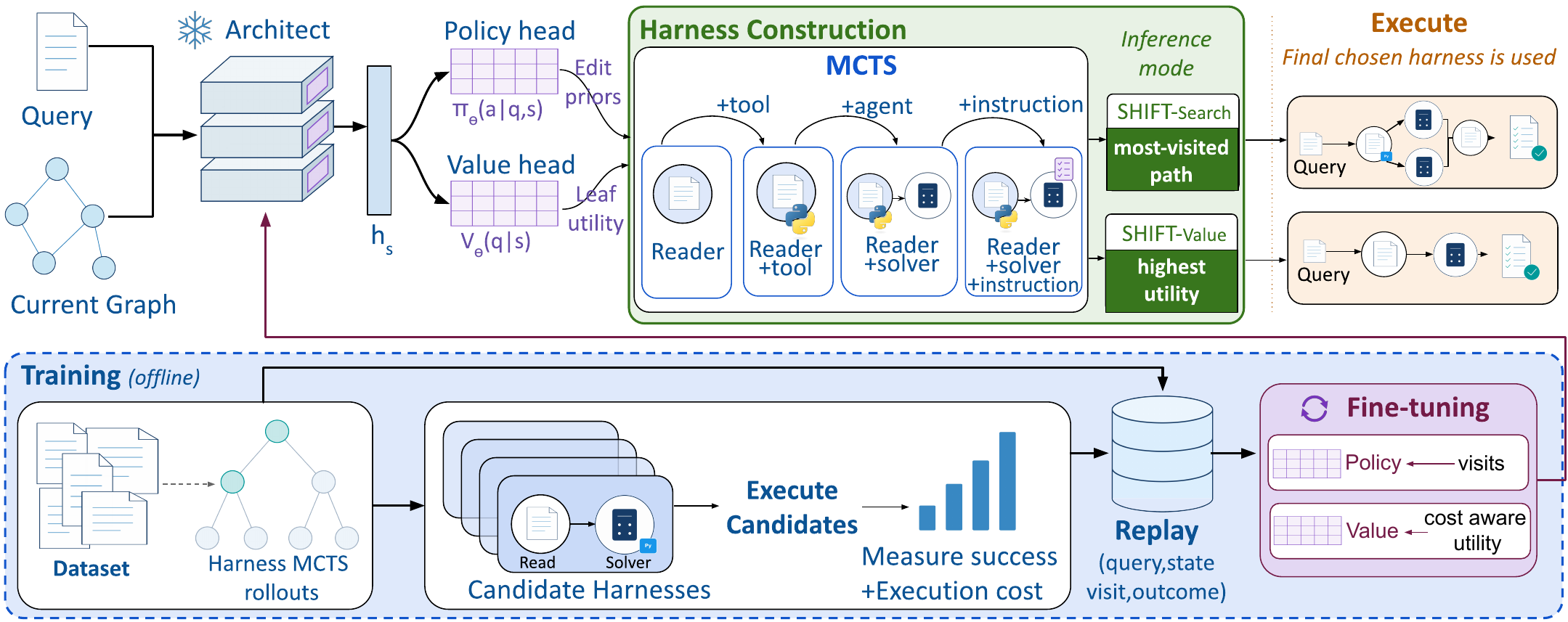}
\caption{\textbf{Overview of SHIFT.} Given a query, the current harness, and its feasible actions, a local Gemma-2-2B architect produces a policy over actions and a utility estimate. MCTS uses these predictions to search over action sequences without executor calls, after which the executor runs only the selected harness. During training, measured outcomes are stored in a replay buffer and used to update the LoRA adapters and policy--value heads.}
    \label{fig:method}
\end{figure}

SHIFT builds a harness for each query by incrementally adding agents, instructions, and tools to an executable graph. Every intermediate graph is runnable, allowing search to compare designs at different construction stages (Section~\ref{sec:edits}). A small local model, the \emph{architect}, proposes actions and predicts harness utility (Section~\ref{sec:architect}). These predictions allow Monte Carlo tree search to explore how successive choices work together without executing candidate designs; only the selected harness is run (Section~\ref{sec:search}). The architect learns these predictions from search decisions and measured execution outcomes during training (Section~\ref{sec:training}). Figure~\ref{fig:method} illustrates the framework, and Algorithm~\ref{alg:search} details the search.

\Needspace{22\baselineskip}
\subsection{Harnesses as Executable Graphs}
\label{sec:edits}

\begin{wrapfigure}{r}{0.43\textwidth}
\vspace{-1em}
\centering
\includegraphics[width=\linewidth]{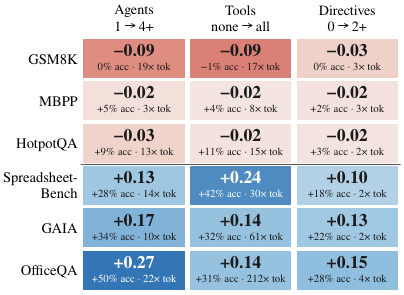}
\caption{\textbf{No single harness fits every task.} Utility change (success minus cost) between harnesses with varying structure; small text: accuracy change and token multiplier. More structure can accompany higher accuracy yet lower utility.}
\label{fig:structure_gain}
\end{wrapfigure}

We represent a harness as a directed graph, because a multi-agent harness is naturally described by who does what and who hands results to whom. Formally, a harness state $s$ is a graph $G=(\mathcal V,\mathcal E)$ with a start node $v_0\in\mathcal V$. Each agent $v\in\mathcal V$ is specified by a role $\rho_v$, an ordered list of directives $D_v$, and a set of permitted tools $T_v\subseteq\mathcal K_d$, where $\mathcal K_d$ is the tool set available in domain $d$. An edge $(u,v)\in\mathcal E$ passes the output of $u$ to $v$. Feedback edges let downstream agents return results for revision. Executing $G$ on a query $q$, starting from $v_0$, produces an answer through the agents' reasoning and tool calls~\cite{yao2022react}. This representation has two advantages. First, it places a single reasoning agent, a retrieval-and-synthesis pipeline, and a workflow with verification loops in one space, so one search procedure covers all of them. Second, it supports explicit checks on graph structure and tool permissions, so search can exclude invalid harnesses.

A harness-building \emph{action} modifies structure, instructions, or tool permissions: it inserts an agent with its edges, adds a feedback edge, appends a directive to some $D_v$, or expands one or more tool sets $T_v$. Adding a Planner, for example, introduces a planning stage, and granting a Reasoner a file reader gives it access to supporting documents. Their benefit varies sharply across tasks (Figure~\ref{fig:structure_gain}). On financial-document questions, harnesses with four or more agents have about 50 percentage points higher accuracy than single-agent harnesses; on arithmetic word problems, they use about 19 times as many execution tokens with little accuracy difference. Components come from a user-defined library, and SHIFT decides for each query how to compose them. Full specifications of role instructions, the initial harness, the action library, and the search settings are given in Appendix~\ref{app:implementation}.

A sequence of at most $H$ actions defines a finite-horizon decision process over harness states. Let $\mathcal S_d$ be the set of executable harnesses that use only tools in $\mathcal K_d$, and $\mathcal A_d$ the action vocabulary. Each action $a_t$ transforms $s_t$ into $s_{t+1}=M(s_t,a_t)$ under a deterministic transition $M$. Feasible actions either modify the harness while preserving executability or stop construction:
\begin{equation}
 \mathcal A(s)=\{a\in\mathcal A_d:M(s,a)\in\mathcal S_d,\ M(s,a)\ne s\}
 \cup\{a_{\mathrm{stop}}\}.
 \label{eq:legal}
\end{equation}
From a minimal harness $s_0\in\mathcal S_d$, a single Coder without tools, the architect chooses actions given the query and the current harness. Construction stops at $a_{\mathrm{stop}}$ or after $H$ actions; since every state is executable, search can compare partial and complete designs.

\begin{algorithm}[tbp]
\caption{Per-query harness construction. The complete procedure is Algorithm~\ref{alg:train} (Appendix~\ref{app:implementation}).}
\label{alg:search}
\small
\begin{algorithmic}[1]
\Require query $q$, architect $(\pi_\theta,V_\theta)$, budget $B$
\State $\mathcal T\gets$ tree rooted at seed harness $s_0$
\For{$b=1,\dots,B$}
  \State descend $\mathcal T$ by Eq.~(\ref{eq:puct}) to leaf $s$
  \State expand $s$ with priors from $\pi_\theta$
  \State back up $V_\theta(q,s)$ \Comment{no executor}
\EndFor
\State \textsc{search}: $s^\star\gets$ most-visited path's end
\State \textsc{value}: $s^\star\gets\arg\max V_\theta$ over candidates
\State \Return $\textsc{Execute}(s^\star,q)$ \Comment{one call}
\end{algorithmic}
\end{algorithm}

\subsection{The Policy--Value Architect}
\label{sec:architect}

A Gemma 4 E2B backbone~\cite{team2026gemma4} reads the query, available tools, feasible actions, and a text rendering of the current graph. Its final-token hidden state is $h_\theta(q,s)$. Two linear heads map it to action logits $\ell_{\theta,a}(q,s)$ and a value logit $\hat v_\theta(q,s)$. Only LoRA adapters~\cite{hu2022lora} and the two heads, collectively $\theta$, are trained; the backbone weights remain frozen. The policy is normalized over feasible actions $\mathcal A(s)$; all others receive zero probability:
\begin{equation}
 \pi_\theta(a\mid q,s)=
 \frac{\mathbf{1}[a\in\mathcal A(s)]\exp(\ell_{\theta,a}(q,s)/T_\pi)}
 {\sum_{b\in\mathcal A(s)}\exp(\ell_{\theta,b}(q,s)/T_\pi)},
 \qquad
 V_\theta(q,s)=\sigma(\hat v_\theta(q,s)).
 \label{eq:policy}
\end{equation}
Here $T_\pi>0$ is the policy temperature and $\sigma$ is the sigmoid. The policy $\pi_\theta$ guides exploration, while $V_\theta(q,s)\in(0,1)$ estimates utility, balancing task success against resource use (Section~\ref{sec:training}). Neither prediction requires running the harness.

\subsection{Searching over Harnesses}
\label{sec:search}

SHIFT searches for a harness with Monte Carlo tree search (MCTS)~\cite{silver2017mastering,hao2023reasoning}, guided by the architect's policy and value. MCTS suits harness design for three reasons. First, the transition $M$ is known and deterministic, so candidate harnesses can be constructed without execution. The architect estimates their utility through local inference, allowing search to compare alternatives before selecting one to run. Second, design choices interact: an agent may become useful only when paired with suitable instructions or tools. MCTS captures these dependencies by evaluating actions through the designs they lead to. Third, search visit counts provide training targets for the policy~\cite{silver2017mastering}, allowing the architect to learn from search and guide future searches toward promising designs.

The search tree is rooted at $s_0$, with harness states as nodes and actions as branches. When search first reaches a node, the architect predicts its policy and utility. The $K_{\mathrm{exp}}$ highest-probability feasible actions, together with $a_{\mathrm{stop}}$, form the expanded set $U(s)$. The policy is then renormalized over $U(s)$ to obtain the search prior $P(a\mid s)$. Each of the $B$ iterations then descends from the root, choosing at each expanded node:
\begin{equation}
 a^\star=\arg\max_{a\in U(s)}\left[
 \widetilde Q(s,a)+c_{\mathrm{puct}}P(a\mid s)
 \frac{\sqrt{\max(1,\widetilde N(s))}}{1+\widetilde N(s,a)}\right],
 \label{eq:puct}
\end{equation}

\begin{wrapfigure}{r}{0.43\textwidth}
\vspace{0.1em}
\centering
\includegraphics[width=\linewidth]{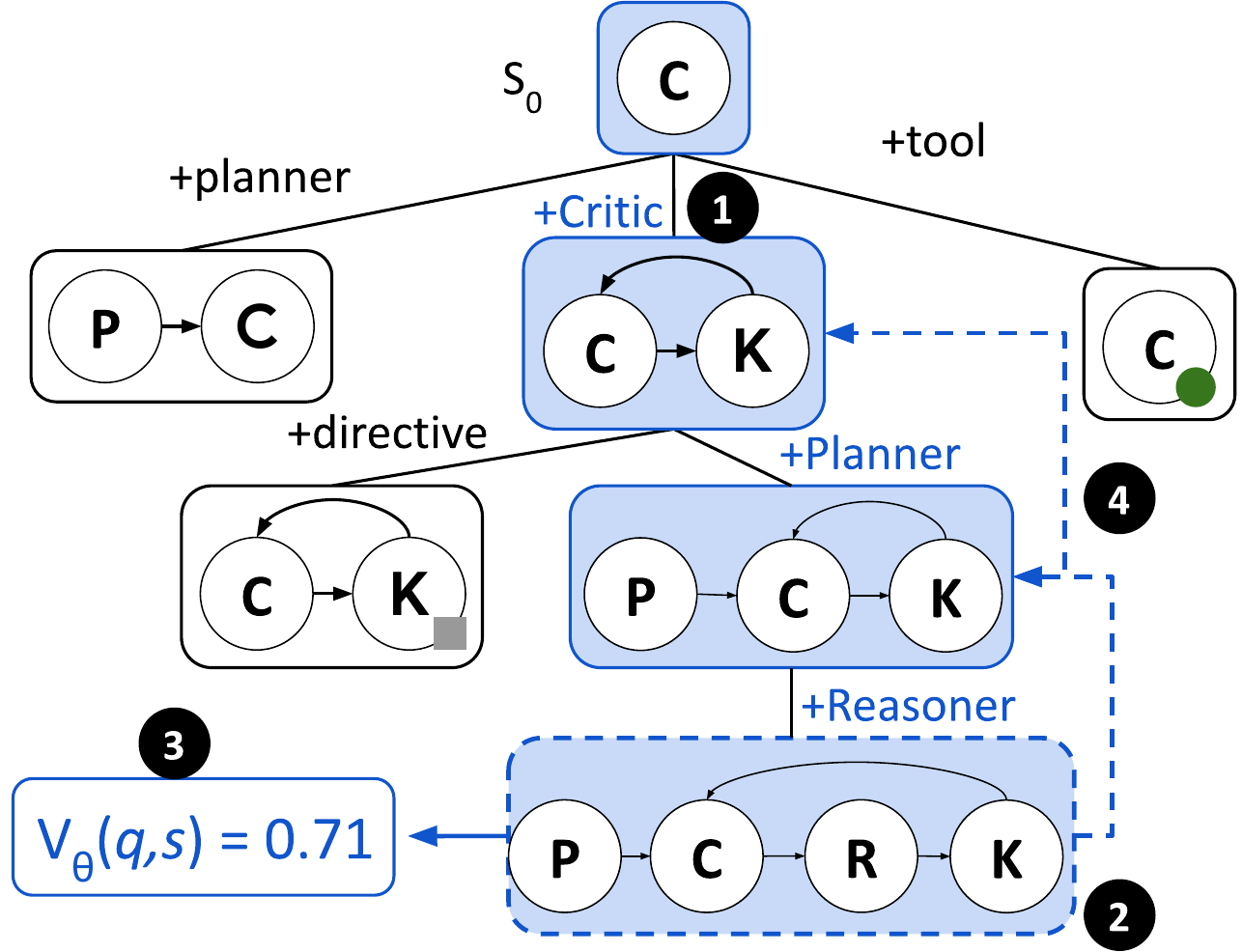}
\caption{One search simulation: (1) select, (2) expand, (3) score with $V_\theta$ (4) back up. C: Coder, P: Planner, R: Reasoner, K: Critic.}
\label{fig:mcts}
\end{wrapfigure}

where $N$ counts visits, $Q(s,a)$ is the mean value backed up through action $a$, and $c_{\mathrm{puct}}>0$ balances the two terms; tildes mark statistics adjusted for parallel search (Appendix~\ref{app:implementation}). The first term favors actions whose continuations have scored well, and the second favors promising actions that have been tried less often. When the descent reaches a new node, selects $a_{\mathrm{stop}}$, or hits depth $H$, the leaf's value is passed back along the path to update $N$ and $Q$. The executor is never called inside this loop.

Once the budget is spent, SHIFT-search executes the harness at the end of the most-visited path, and SHIFT-value executes the candidate from the most-visited branches with the highest predicted utility, searching again with a larger budget if that value is low (Appendix~\ref{app:decoding}).

\textbf{Amortized design cost.} SHIFT replaces executor calls with local inference. Training is paid once and amortized over later queries, whereas search runs for every query and pays off when it makes execution cheaper: search matches greedy construction in accuracy at about 30\% lower executor-token cost (Section~\ref{sec:cost}).

\subsection{Learning from Execution}
\label{sec:training}

The architect learns its policy from search and its value estimates from executed harnesses within a shared training loop (Algorithm~\ref{alg:train}). For each training query in $\mathcal Q_d$, SHIFT runs search with exploration noise at the root, samples a construction path according to visit counts, and executes the resulting harness and up to $K_{\mathrm{exec}}$ other candidates from the tree.

\textbf{Cost-aware reward.} Each executed harness receives a reward that credits task success and subtracts penalties for execution tokens, latency, timeouts, and additional tools and agents, each scaled by a domain-specific reference. The reward is clipped and rescaled to a target $z\in[0,1]$, the utility the value head learns to predict (Equation~\ref{eq:reward}, Appendix~\ref{app:implementation}).

\textbf{Training targets.} Each training example $i$ pairs a query $q_i$ with a harness state $s_i$. If search visited children of $s_i$, its policy target is the share of visits each expanded action received, $p_i(a)=N(s_i,a)/\sum_{b\in U(s_i)}N(s_i,b)$ for $a\in U(s_i)$. If $s_i$ was executed, its value target is its measured utility $z_i$. Intermediate states thus never inherit another harness's outcome.

\textbf{Online objective.} Examples are drawn from a replay buffer in minibatches $\mathcal I$ that favor informative examples, with importance weights $w_i>0$ correcting for this bias. Let $\mathcal I_\pi\subseteq\mathcal I$ and $\mathcal I_V\subseteq\mathcal I$ be the examples with policy and value targets. The architect minimizes
\begin{equation}
 \mathcal L_{\mathrm{online}}=\frac{1}{|\mathcal I|}\Big[
 \sum_{i\in\mathcal I_\pi}w_i\big(\operatorname{CE}(p_i,\pi_i^U)-\beta\,\mathcal H(\pi_i)\big)
 +\lambda_v\sum_{i\in\mathcal I_V}w_i\operatorname{BCE}(z_i,V_i)\Big].
 \label{eq:online}
\end{equation}
The first sum trains the policy $\pi_i=\pi_\theta(\cdot\mid q_i,s_i)$ to match the search visit distribution $p_i$. Cross-entropy is computed using $\pi_i^U$, the policy renormalized over expanded actions $U(s_i)$, so unexamined actions do not count as negative evidence. The entropy term $\mathcal H(\pi_i)$, weighted by $\beta\ge0$, encourages exploration. The second sum trains $V_i=V_\theta(q_i,s_i)$ to predict measured utility $z_i$ using binary cross-entropy, weighted by $\lambda_v\ge0$.

\textbf{Ranking objective.} Search needs to distinguish better harnesses from worse ones for the same query. At the end of each epoch, we therefore add a ranking loss that trains the value head to preserve their measured utility ordering. Let $\mathcal J$ be a sample of pairs $(i,j)$ executed on the same query whose utilities differ by at least $\delta>0$, and let $\hat v_i=\hat v_\theta(q_i,s_i)$ be the value logit before the sigmoid. Then \begin{equation}
 \mathcal L_{\mathrm{rank}}=
 \frac{1}{|\mathcal J|}\sum_{(i,j)\in\mathcal J}
 \max\{0,\ m-\operatorname{sgn}(z_i-z_j)(\hat v_i-\hat v_j)\},
 \label{eq:rank}
\end{equation}
where $\operatorname{sgn}(z_i-z_j)$ is $+1$ if harness $i$ scored higher and $-1$ otherwise. The loss is zero once the better harness leads by at least the margin $m>0$ in logit space and grows linearly otherwise; it is added to the value cross-entropy with weight $\lambda_r\ge0$.

%% file: sections/experiment.tex
\section{Experimental Results}
\label{sec:experiments}

We evaluate SHIFT on six benchmarks that range from short arithmetic problems to long, tool-heavy document and assistant tasks. Section~\ref{sec:main} compares the accuracy of SHIFT's harnesses with fixed and learned workflow designs, Section~\ref{sec:cost} examines how search trades accuracy against execution cost, Section~\ref{sec:controls} identifies the components that account for the gains, and Section~\ref{sec:executor_transfer} tests transfer to harder tasks without retraining.

\subsection{Evaluation Setup}
\label{sec:setup}

\textbf{Benchmarks and baselines.} We evaluate mathematical reasoning (GSM8K~\cite{cobbe2021gsm8k}), multi-hop question answering (HotpotQA~\cite{yang2018hotpotqa}), code generation (MBPP~\cite{austin2021mbpp}), spreadsheet manipulation (SpreadsheetBench~\cite{ma2024spreadsheetbench}), document-grounded question answering (OfficeQA Pro~\cite{opsahlong2026officeqa}), and general-purpose assistance (GAIA~\cite{mialon2024gaia}), for a total of 9,193 evaluation tasks. We use official splits where they exist and study-specific partitions for SpreadsheetBench, OfficeQA, and GAIA (Appendix~\ref{app:reporting}). We compare against 17 baselines in the three groups of Section~\ref{sec:related}: fixed harnesses, spanning prompting strategies~\cite{wei2022chain,wang2023selfconsistency,wang2023planandsolve,madaan2023selfrefine,yao2023tree}, tool-using and multi-agent frameworks~\cite{yao2022react,chen2024agentverse}, and DSPy programs with access to every tool, unoptimized or optimized with MIPROv2 and GEPA~\cite{khattab2023dspy,opsahl2024mipro,agrawal2025gepa}; workflow optimizers~\cite{zhang2025aflow,zhuge2024language,cheng2024trace}; and query-adaptive composition~\cite{liu2024dylan}. All methods use Gemini 3.5 Flash~\cite{team2023gemini} as the executor unless otherwise specified. Appendix~\ref{app:baseline_provenance} details each implementation.

\textbf{Evaluation metrics.} We report accuracy as the mean and sample standard deviation over three inference runs. The mean weights all six benchmarks equally. Execution cost is the geometric mean over benchmarks of mean executor tokens relative to Direct; it excludes architect inference and training, whose cost we report separately. Architect and search settings are given in Appendix~\ref{app:implementation}.

\subsection{RQ1: Does Query-Specific Construction Improve Accuracy?}
\label{sec:main}

\textbf{SHIFT constructs the most accurate harnesses.} SHIFT-search attains the highest six-benchmark mean accuracy in Table~\ref{tab:main}, at 79.9\%, an improvement of 7.2\% over the strongest baseline, Trace (72.7\%); a paired task-bootstrap 95\% confidence interval runs from 4.42\% to 9.88\% (Appendix~\ref{app:paired_main}). The gains concentrate on the two long-horizon, tool-intensive benchmarks, where no baseline is strong: every baseline stays below 30\% on OfficeQA, yet SHIFT-search reaches 59.3\% there and 68.6\% on GAIA, exceeding Trace by 30.0\% and 15.0\%. SHIFT-value also outperforms every baseline on both, at 39.0\% and 64.7\%. These gains are broad rather than driven by a few tasks: against Trace, search wins on 24 OfficeQA questions and loses on four ($p=2\times10^{-4}$, paired sign test), and wins on 19 GAIA questions and loses on seven ($p=0.03$) (Appendix~\ref{app:paired_main}).

\input{sections/table1}
\begin{figure}[tbp]
\centering
\begin{minipage}[t]{0.485\textwidth}
\centering\vspace{0pt}
\includegraphics[width=\linewidth]{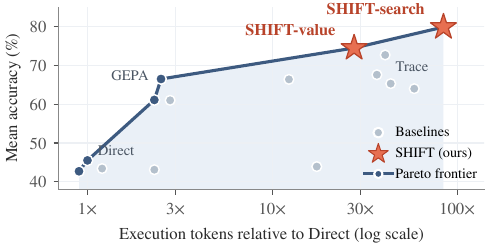}
\caption{Accuracy--cost Pareto frontier for the six-benchmark methods in Table~\ref{tab:main}. Both SHIFT variants lie on the frontier.}
\label{fig:frontier}
\end{minipage}\hfill
\begin{minipage}[t]{0.485\textwidth}
\centering\vspace{0pt}
\includegraphics[width=\linewidth]{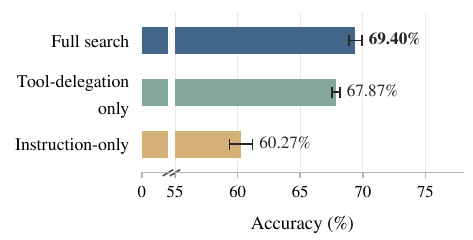}
\caption{HotpotQA accuracy (500 questions; bars show SD over three runs). The broken axis omits 3--55\%.}
\label{fig:rq1_edits}
\end{minipage}
\end{figure}

\textbf{Search pays off when two conditions hold:} harness structure strongly affects success, and no existing design has already found the right one. OfficeQA and GAIA meet both: in training, adding agents or tools raises success by 50\% and 34\% (Figure~\ref{fig:structure_gain}), yet the best baseline reaches only 29.3\% and 53.6\%. Otherwise, SHIFT stays close to the leader. On GSM8K, HotpotQA, and MBPP, a single agent nearly matches the best method (16, 13, and 6 of the 17 baselines lie within 2\%), so the executor rather than the harness sets the ceiling, and SHIFT-search is within 1.2\% of it; without OfficeQA, SHIFT-search still has the highest mean (84.0\% vs.\ 81.4\% for Trace).

\textbf{Ablation study on the action space.} To test whether the gains require choosing agents, instructions, and tools together, we restrict SHIFT's action space at inference time on 500 HotpotQA questions with three runs each (Figure~\ref{fig:rq1_edits}). Full search reaches 69.4\%, compared with 67.9\% when the architect may only delegate tools and 60.3\% when it may only add instructions. Instructions tell an agent what to do, tools determine what it can do, and additional agents decide who does it; SHIFT gains from choosing these together rather than any one alone. The restricted variants start from a single Coder and cannot add roles, so the comparison measures the full construction space (Appendix~\ref{app:edit_spaces}).

\begin{table}[tbp]
\centering\scriptsize\setlength{\tabcolsep}{2pt}
\caption{Decoding comparison using the same trained architect within each benchmark. Cells report accuracy (\%, mean $\pm$ sample standard deviation over three evaluation runs). Mean weights all six benchmarks equally; Cost is the geometric mean of execution-token ratios to Direct.}
\label{tab:shift_modes}
\begin{tabularx}{\textwidth}{@{}l*{8}{Y}@{}}
\toprule
Mode & GSM8K & Hotpot & MBPP & SpreadSheet. & OfficeQA & GAIA & Mean & Cost\\
\midrule
SHIFT-greedy & $98.0_{\pm 0.9}$ & $69.2_{\pm 0.1}$ & $96.4_{\pm 1.0}$ & $87.5_{\pm 2.5}$ & $52.8_{\pm 2.8}$ & $75.2_{\pm 15.0}$ & 79.8 & 120.2\\
SHIFT-policy-veto & $98.1_{\pm 0.5}$ & $69.2_{\pm 0.0}$ & $96.4_{\pm 0.6}$ & $85.6_{\pm 1.3}$ & $59.3_{\pm 9.9}$ & $66.0_{\pm 4.9}$ & 79.1 & 115.0\\
SHIFT-value-greedy & $98.0_{\pm 0.1}$ & $64.6_{\pm 0.2}$ & $95.6_{\pm 0.6}$ & $58.6_{\pm 1.3}$ & $10.6_{\pm 3.7}$ & $34.0_{\pm 3.0}$ & 60.2 & 5.3\\
\midrule
SHIFT-search & $98.2_{\pm 0.5}$ & $68.7_{\pm 0.1}$ & $96.6_{\pm 0.2}$ & $87.8_{\pm 1.7}$ & $59.3_{\pm 5.1}$ & $68.6_{\pm 5.2}$ & 79.9 & 84.3\\
SHIFT-value & $97.7_{\pm 0.1}$ & $67.5_{\pm 0.2}$ & $96.1_{\pm 0.8}$ & $81.7_{\pm 0.8}$ & $39.0_{\pm 6.5}$ & $64.7_{\pm 7.1}$ & 74.5 & 27.7\\
\bottomrule
\end{tabularx}
\end{table}

\subsection{RQ2: How Does Harness Selection Balance Accuracy and Cost?}
\label{sec:cost}

\begin{wrapfigure}[12]{r}{0.43\textwidth}
\centering
\vspace{-2em}
\includegraphics[width=\linewidth]{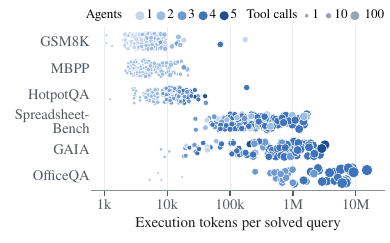}

\caption{Solved test queries of SHIFT-value: execution tokens, agents in the harness (shade), and tool calls (size).}
\label{fig:tokens_per_query}
\end{wrapfigure}

\textbf{A cheaper operating point.} SHIFT-value selects by predicted utility, which rewards cheaper harnesses, and sizes each harness to the query (Figure~\ref{fig:tokens_per_query}): it solves arithmetic and coding questions with one or two agents and few tool calls, and long-horizon tasks with larger harnesses that call tools extensively. It reaches 74.5\%, above every baseline, while using 32\% fewer execution tokens than Trace and one third those of SHIFT-search. Figure~\ref{fig:frontier} places both variants on the accuracy--cost Pareto frontier: no baseline is both more accurate and cheaper than either. Against Trace, SHIFT-value is also more accurate on OfficeQA (+9.7\%) and GAIA (+11.1\%). One architect thus yields both the most accurate harness and a much cheaper one.

\begin{figure}[t]
\centering

\begin{minipage}[t]{0.49\textwidth}
\vspace{0pt}
\centering
\small
\setlength{\tabcolsep}{3pt}
\renewcommand{\arraystretch}{1.08}
\captionof{table}{Accuracy (\%) of the harness chosen from identical candidate pools by the learned value $V_\theta$ (400 HotpotQA and 200 GSM8K questions, Gemma 3 27B).}
\label{tab:rq3_selection}
\begin{tabular*}{\linewidth}{@{\extracolsep{\fill}}lcc@{}}
\toprule
Selection rule & HotpotQA & GSM8K\\
\midrule
Lowest $V_\theta$ & $35.6_{\pm 1.2}$ & $86.1_{\pm 0.2}$\\
Uniform & $39.1_{\pm 0.3}$ & $89.7_{\pm 0.7}$\\
Second-highest $V_\theta$ & $37.7_{\pm 1.1}$ & $89.9_{\pm 0.5}$\\
\midrule
\textbf{Highest $V_\theta$ (ours)} & $\mathbf{49.6}_{\pm 0.4}$ & $\mathbf{92.6}_{\pm 0.9}$\\
\bottomrule
\end{tabular*}
\end{minipage}
\hfill
\begin{minipage}[t]{0.49\textwidth}
\vspace{0pt}
\centering

\includegraphics[width=\linewidth]{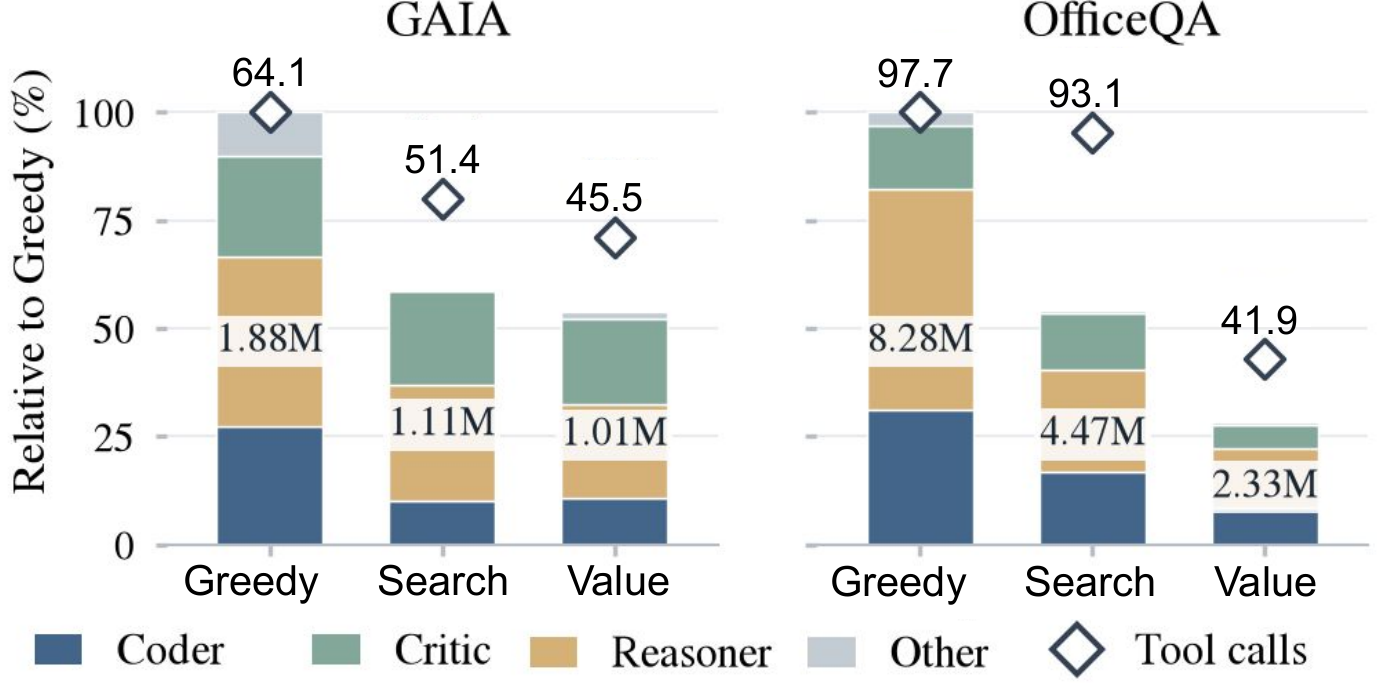}

\captionof{figure}{Tokens, tool calls, relative to Greedy. Labels mean tokens/task in millions inside bars and mean tool calls/task above diamonds.}
\label{fig:role_tokens}
\end{minipage}
\end{figure}

\begin{figure}[t]
\centering
\begin{minipage}[t]{0.48\textwidth}
\vspace{0pt}\centering
\vspace{-4em}
\includegraphics[width=\linewidth]{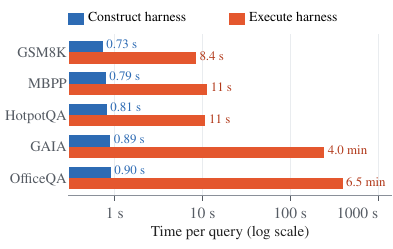}
\caption{Time per query to construct a harness (one H100, 64 simulations) and to execute it (median over test queries, SHIFT-value).}
\label{fig:construct_time}
\end{minipage}\hfill
\begin{minipage}[t]{0.49\textwidth}
\vspace{0pt}\centering\small
\setlength{\tabcolsep}{3pt}
\captionof{table}{MATH-500 accuracy (\%) by difficulty level, using the architect trained on GSM8K without retraining (Llama 4 Scout).}
\label{tab:math_levels}
\begin{tabularx}{\linewidth}{@{}Xrrr@{}}
\toprule
Configuration & Lv.~1--3 & Lv.~4--5 & All\\
\midrule
Fixed single Coder & 92.4 & 71.0 & 81.2\\
Uniform $\pi$, constant $V$ & 92.0 & 72.9 & 82.0\\
Learned $\pi$, constant $V$ & 91.6 & 74.0 & 82.4\\
Uniform $\pi$, learned $V$ & 91.6 & 71.8 & 81.2\\
\midrule
SHIFT (learned $\pi$ and $V$) & \textbf{92.4} & \textbf{75.6} & \textbf{83.6}\\
\bottomrule
\end{tabularx}
\end{minipage}
\end{figure}

\textbf{Search lowers execution cost without losing accuracy.} To isolate search, we decode harnesses from the same trained architect in three simpler ways (Table~\ref{tab:shift_modes}): \emph{greedy} construction applies the policy's most probable action at each step, \emph{policy-veto} lets the value head reject low-utility policy proposals, and \emph{value-greedy} moves to the one-action change the value head scores highest. Search matches greedy construction in mean accuracy (79.9\% versus 79.8\%) while using about 30\% fewer execution tokens, and 27\% fewer than policy-veto. Greedy decoding is itself a product of search: the policy is trained to reproduce search's visit counts (Section~\ref{sec:training}), so it inherits search's accuracy, while search at inference adds lookahead on predicted utility, which charges for execution cost. The search itself adds little: constructing a harness takes under a second, a small fraction of the time needed to execute it (Figure~\ref{fig:construct_time}). The savings are not simply fewer tool calls: on OfficeQA, tokens fall by 46\% while tool calls barely change (Figure~\ref{fig:role_tokens}), and search is cheaper on 60--98\% of questions. Value alone is not enough: value-greedy is the cheapest strategy but falls to 60.2\%, since one-step lookahead cannot credit an agent whose benefit appears only after later actions (Appendix~\ref{app:decoding}).


\subsection{RQ3: How Effectively Does Learned Value Guide Harness Selection?}
\label{sec:controls}

\begin{wrapfigure}[12]{r}{0.43\textwidth}
\centering
\includegraphics[width=\linewidth]{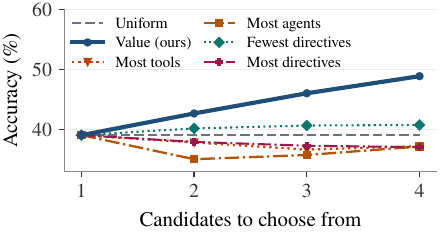}
\caption{HotpotQA accuracy when choosing from $k$ candidates.}
\label{fig:offline_ranking}
\end{wrapfigure}

We compare maximum-value and uniform selection over identical candidate pools, with values frozen before execution (Gemma 3 27B executor). Value selection raises accuracy by 10.5\% on HotpotQA and 2.9\% on GSM8K with half and one fifth of the execution tokens, respectively (Table~\ref{tab:rq3_selection}; Appendix~\ref{app:ranking_protocol}). Its top choice beats the second-ranked candidate by 11.9\% on HotpotQA, and the benefit grows as more candidates are considered (Figure~\ref{fig:offline_ranking}). Fixed structural recipes fail: always picking the most agents, tools, or directives is worse than random selection, and picking the fewest directives barely beats it. Picking the fewest agents helps on HotpotQA but fails on long-horizon tasks (OfficeQA: 9.4\% vs.\ 59.3\% with four or more agents; Table~\ref{tab:structure_gain}).

\textbf{Accurate and inexpensive choices.} Because the value head is trained on a utility that rewards success and charges for cost, it learns to prefer the least expensive harness that still solves the query. This is what SHIFT-value exploits: it recovers several candidates from the search tree and executes the one the value head ranks highest, reaching the accuracy of the strongest baselines at a fraction of the cost of SHIFT-search (Section~\ref{sec:cost}).

\subsection{RQ4: Does the Architect Transfer to Harder Tasks?}
\label{sec:executor_transfer}

An architect trained only on GSM8K transfers to the 500 MATH-500 problems (Llama 4~\cite{meta2025llama4scout} executor), where SHIFT reaches 83.6\% against 81.2\% for a fixed single Coder. The gain concentrates on the hardest problems (Table~\ref{tab:math_levels}): on levels 4--5, SHIFT reaches 75.6\% against 71.0\%, and both its learned policy and its learned value contribute (Appendix~\ref{app:math_transfer}).

%% file: sections/table1.tex
\begin{table}[tbp]
\centering
\caption{Accuracy (\%, mean and standard deviation over three runs) with a shared Gemini 3.5 Flash executor.
The first three benchmarks are short-horizon tasks that a single agent nearly solves; the last three are long-horizon and tool-intensive. Cost is the geometric mean of execution tokens relative to Direct. \textbf{Bold}: best; \underline{underline}: second best.
$^{\ddagger}$Cost average excludes SpreadsheetBench: these methods answer in text, so they score zero there by construction and were not run (Appendix~\ref{app:baseline_provenance}).}
\label{tab:main}
\footnotesize
\setlength{\tabcolsep}{1.5pt}
\renewcommand{\arraystretch}{1.12}
\begin{tabularx}{\textwidth}{@{}l>{\centering\arraybackslash\hsize=1.0\hsize}X>{\centering\arraybackslash\hsize=1.1\hsize}X>{\centering\arraybackslash\hsize=1.0\hsize}X>{\centering\arraybackslash\hsize=1.14\hsize}X>{\centering\arraybackslash\hsize=1.1\hsize}X>{\centering\arraybackslash\hsize=1.1\hsize}X!{\color{gray!45}\vrule width 0.4pt}>{\centering\arraybackslash\hsize=0.78\hsize}X>{\centering\arraybackslash\hsize=0.78\hsize}X@{}}
\toprule
& \multicolumn{3}{c}{Short-horizon} & \multicolumn{3}{c!{\color{gray!45}\vrule width 0.4pt}}{Long-horizon, tool-intensive} & \multicolumn{2}{c}{\textbf{Aggregate}}\\
\cmidrule(lr){2-4}\cmidrule(lr){5-7}\cmidrule(l){8-9}
Method & GSM8K & HotpotQA & MBPP & Sheet & OfficeQA & GAIA & \textbf{Mean}$\,\uparrow$ & \textbf{Cost}$\,\downarrow$\\
\midrule
\multicolumn{7}{@{}l!{\color{gray!45}\vrule width 0.4pt}}{\textit{\color{gray}Prompting}} & & \\
Direct & 97.7$\pm${0.1} & 67.9$\pm${0.1} & 93.6$\pm${0.6} & 0.0 & 8.9$\pm${3.7} & 4.6$\pm${1.1} & 45.5 & 1.0 \\
CoT & 97.4$\pm${0.3} & 68.4$\pm${0.1} & 78.3$\pm${0.6} & 0.0 & 4.1$\pm${1.4} & 3.3$\pm${1.1} & 41.9 & 0.9$^{\ddagger}$ \\
CoT (few-shot) & 97.6$\pm${0.1} & 68.4$\pm${0.2} & 82.6$\pm${2.2} & 0.0 & 4.9$\pm${2.4} & 2.6$\pm${1.1} & 42.7 & 0.9 \\
Self-Consistency & 97.7$\pm${0.1} & 68.5$\pm${0.0} & 90.0$\pm${0.4} & 0.0 & 6.5$\pm${2.8} & 6.5$\pm${1.1} & 44.9 & 5.2$^{\ddagger}$ \\
Plan-and-Solve & 97.4$\pm${0.2} & 68.0$\pm${0.1} & 85.5$\pm${3.9} & 0.0 & 0.8$\pm${1.4} & 4.6$\pm${1.1} & 42.7 & 1.9$^{\ddagger}$ \\
Self-Refine & 97.8$\pm${0.2} & 68.5$\pm${0.0} & 86.3$\pm${1.6} & 0.0 & 3.3$\pm${2.8} & 2.6$\pm${1.1} & 43.1 & 2.3 \\
Tree-of-Thoughts & 97.8$\pm${0.1} & 68.3$\pm${0.0} & 91.6$\pm${0.2} & 0.0 & 0.8$\pm${1.4} & 5.2$\pm${2.3} & 43.9 & 17.4 \\

\multicolumn{7}{@{}l!{\color{gray!45}\vrule width 0.4pt}}{\textit{\color{gray}Tool use and multi-agent}} & & \\
ReAct & \underline{98.2}$\pm${0.7} & 69.1$\pm${1.1} & \textbf{97.1}$\pm${0.6} & 84.4$\pm${5.5} & 16.3$\pm${3.7} & 33.3$\pm${7.1} & 66.4 & 12.3 \\
LAP & 95.5$\pm${0.4} & 17.0$\pm${0.3} & 7.3$\pm${0.8} & 6.1$\pm${4.3} & 1.6$\pm${1.4} & 6.5$\pm${3.0} & 22.3 & 7.8 \\
AgentVerse & 97.3$\pm${0.2} & 68.2$\pm${0.1} & 96.1$\pm${0.0} & 87.8$\pm${1.3} & 4.9$\pm${2.4} & 51.0$\pm${5.2} & 67.6 & 36.8 \\

\multicolumn{7}{@{}l!{\color{gray!45}\vrule width 0.4pt}}{\textit{\color{gray}DSPy programs}} & & \\
DSPy (unoptimized) & 97.6$\pm${0.1} & 68.1$\pm${0.1} & 94.2$\pm${0.4} & 75.8$\pm${3.3} & 9.8$\pm${0.0} & 20.9$\pm${3.0} & 61.1 & 2.3 \\
DSPy-MIPROv2 & 97.6$\pm${0.0} & 60.5$\pm${0.6} & 94.2$\pm${0.4} & 78.1$\pm${7.0} & 8.1$\pm${1.4} & 27.5$\pm${2.0} & 61.0 & 2.8 \\
DSPy-GEPA & 97.7$\pm${0.2} & \underline{69.6}$\pm${0.1} & 95.3$\pm${0.4} & \textbf{95.3}$\pm${1.7} & 11.4$\pm${2.8} & 29.4$\pm${2.0} & 66.5 & 2.5 \\

\multicolumn{7}{@{}l!{\color{gray!45}\vrule width 0.4pt}}{\textit{\color{gray}Workflow optimization and query-adaptive}} & & \\
AFlow & 97.4$\pm${0.1} & 68.2$\pm${0.1} & 81.7$\pm${0.7} & 0.0 & 1.6$\pm${1.4} & 11.8$\pm${2.0} & 43.4 & 1.2 \\
GPTSwarm & 97.9$\pm${0.2} & 64.1$\pm${7.1} & 95.7$\pm${0.7} & 83.9$\pm${2.9} & 9.8$\pm${3.4} & 32.7$\pm${3.0} & 64.0 & 58.6 \\
Trace & \textbf{98.4}$\pm${0.2} & \textbf{69.9}$\pm${1.2} & 96.6$\pm${0.2} & \underline{88.6}$\pm${3.5} & 29.3$\pm${11.2} & 53.6$\pm${12.6} & 72.7 & 40.8 \\
DyLAN & 98.1$\pm${0.2} & 64.7$\pm${2.8} & \underline{97.0}$\pm${0.2} & 88.1$\pm${2.9} & 12.2$\pm${2.4} & 32.0$\pm${1.1} & 65.3 & 43.7 \\
\midrule
\rowcolor{shiftrow}\textbf{SHIFT-value} & 97.7$\pm${0.1} & 67.5$\pm${0.2} & 96.1$\pm${0.8} & 81.7$\pm${0.8} & \underline{39.0}$\pm${6.5} & \underline{64.7}$\pm${7.1} & \underline{74.5} & 27.7 \\
\rowcolor{shiftrow}\textbf{SHIFT-search} & \underline{98.2}$\pm${0.5} & 68.7$\pm${0.1} & 96.6$\pm${0.2} & 87.8$\pm${1.7} & \textbf{59.3}$\pm${5.1} & \textbf{68.6}$\pm${5.2} & \textbf{79.9} & 84.3 \\
\bottomrule
\end{tabularx}
\end{table}

%% file: sections/conclusion.tex
\Needspace{18\baselineskip}
\section{Conclusion}

We presented SHIFT, which casts harness design as search over executable graphs and learns, from executed harnesses, to predict which designs will succeed and at what cost. By moving execution out of the search loop, SHIFT constructs a harness for each query with a small local model in under a second on one GPU and executes only the selected harness. Across six benchmarks, it builds more accurate harnesses than 17 baselines on a shared executor, with the largest gains on long, tool-intensive tasks, and a value-based variant attains a higher mean accuracy than every baseline at lower cost than the strongest. These gains stem from choosing agents, instructions, and tools jointly, concentrate where structure matters and existing designs fall short, and carry over to harder problems without retraining. SHIFT shows how harness design can be learned from execution and adapted to individual queries. Training separate architects per benchmark incurs an upfront compute cost. Future work could develop an architect that generalizes across datasets and executors, reducing repeated training, and improve value estimation using outcomes averaged over executions.

%% file: sections/appendix.tex
\clearpage
\appendix

\section*{Appendix}

The appendix is organized in three parts. Appendix~\ref{app:implementation} specifies SHIFT's action space, architect, search, reward, training, and training cost, and Appendix~\ref{app:decoding} compares the ways of decoding a harness from the trained architect, shows how cost and harness size vary across queries, and measures architect overhead. Appendices~\ref{app:reporting} and~\ref{app:baseline_provenance} describe the benchmarks, evaluation protocol, and baselines. Appendices~\ref{app:rq1}, \ref{app:rq3}, and~\ref{app:rq4} give further results for Sections~\ref{sec:main}, \ref{sec:controls}, and~\ref{sec:executor_transfer}.

\section{Method and Implementation Details}
\label{app:implementation}

This appendix specifies the components that Sections~\ref{sec:edits}--\ref{sec:training} describe: the action space over harnesses, the architect's observation, the batched form of the tree search, the execution reward that defines the architect's learning target, and the training procedure.

\subsection{Action space}

\begin{table}[tbp]
\centering\small
\caption{Search settings shared by all benchmarks.}
\label{tab:search_bounds}
\begin{tabular}{lr}
\toprule
Setting & Value\\
\midrule
Actions in the vocabulary & 40\\
Top actions per node, $K_{\mathrm{exp}}$ (stop always retained) & 10\\
Maximum number of actions, $H$ & 10\\
Search simulations, $B$ & 64\\
Exploration constant, $c_{\mathrm{puct}}$ & 2.5\\
Policy temperature, $T_\pi$ & 1\\
Leaves evaluated per architect forward pass & 8\\
Architect input length & 2,048 tokens\\
\midrule
SHIFT-value: candidates ranked & 5\\
SHIFT-value: re-search below value / extra simulations & 0.55 / 192\\
Policy-veto: proposals checked / value threshold & 4 / 0.55\\
\bottomrule
\end{tabular}
\end{table}

Every search begins from the same minimal harness $s_0$, a single Coder agent without tools, so that any additional structure must be justified by search rather than assumed. From $s_0$, SHIFT grows a harness through a vocabulary of 40 actions (Table~\ref{tab:action_catalogue}) that mirror the three components of a harness in Section~\ref{sec:edits}: ten structural actions that add an agent or a feedback loop, grant the full tool set, or stop; fifteen directives that append an instruction to an existing agent; and fifteen tool grants that give an existing agent a single tool. The feasible set $\mathcal A(s)$ contains only actions that change the harness and keep it executable: directives and tool grants require their target agent to exist, and a tool can be granted only if the benchmark provides it. Because every action changes a single component, a path through the search tree records exactly which decisions produced the final harness. The library is shared across benchmarks: every structural action and directive is available on every benchmark, and only tool grants depend on the tools a benchmark provides, so the architect must learn from execution which components help where. The resulting design space is large: counting every combination of at most ten actions that respects these dependencies gives about $1.5\times10^{7}$ distinct harnesses, somewhat fewer on benchmarks that offer fewer tools. Table~\ref{tab:search_bounds} summarizes the search configuration used in all experiments.

\begin{table}[tbp]
\centering\small
\caption{The action vocabulary. Directives are appended at most once to their target agent.}
\label{tab:action_catalogue}
\begin{tabularx}{\textwidth}{@{}rlX@{}}
\toprule
ID & Target & Action\\
\midrule
1--6 & Graph & Add a Planner, Researcher, Reasoner, Critic, Verifier, or Aggregator agent.\\
7--8 & Graph & Add a feedback loop from the Verifier or the Critic back to the executing agent.\\
9 & Graph & Give every executing agent the full tool set.\\
10 & Graph & Stop and return the current harness.\\
\midrule
11 & Coder & Think step-by-step before generating code.\\
12 & Coder & Write clean, concise code without extra comments.\\
13 & Coder & Ensure all spreadsheet formula generation conforms strictly to Excel standards.\\
14 & Coder & Include error handling for division by zero and empty datasets.\\
15 & Reasoner & Think step-by-step and show your reasoning path.\\
16 & Reasoner & Provide a direct, concise analytical answer.\\
17 & Critic & Explicitly check units (e.g., millions vs.\ billions) and bounds.\\
18 & Verifier & Verify outputs using code execution whenever possible.\\
19 & Planner & Decompose the query into detailed, granular sub-tasks.\\
20 & Researcher & Align table row and column headers carefully before aggregating values.\\
21 & Aggregator & Provide a high-level summary followed by the final answer.\\
22 & Aggregator & Output the final answer only, without any prefix.\\
23 & Reasoner & Write out symbolic formulas and substitute variables step-by-step.\\
24 & Critic & Check that the answer directly addresses the query in the required format.\\
25 & Verifier & Write assertions that check the output's type, bounds, and format.\\
\midrule
26--30 & Coder & Grant web search, web scraper, file reader, file writer, or calculator.\\
31--34 & Reasoner & Grant web search, web scraper, file reader, or calculator.\\
35--36 & Researcher & Grant web search or web scraper.\\
37--38 & Critic & Grant file reader or file writer.\\
39--40 & Verifier & Grant Python execution or calculator.\\
\bottomrule
\end{tabularx}
\end{table}

\textbf{Tool availability.} GSM8K, HotpotQA, OfficeQA, and SpreadsheetBench provide a file reader, file writer, Python execution, and a calculator. MBPP provides the first three, and GAIA provides all four together with web search and a web scraper. When an agent is added, it receives a small default tool set that fits its role, restricted to the tools the benchmark provides: Researchers and Critics get the file reader, Reasoners the calculator, and Verifiers Python execution and the calculator.

\subsection{Architect, search, and reward}

\begin{table}[tbp]
\centering\small
\caption{Benchmark-specific reference scales for the token and latency penalties (Equation~\ref{eq:reward}).}
\label{tab:reward_scales}
\begin{tabular}{lrr}
\toprule
Benchmark & Tokens $\rho_t$ & Latency $\rho_\ell$ (s)\\
\midrule
GSM8K & 30,000 & 25\\
HotpotQA & 40,000 & 30\\
MBPP & 60,000 & 60\\
SpreadsheetBench & 250,000 & 100\\
OfficeQA & 1,500,000 & 400\\
GAIA & 300,000 & 200\\
\bottomrule
\end{tabular}
\end{table}

\textbf{Observation.} The architect conditions on the information that determines which action is useful: a short description of the benchmark, its available tools, the query, the feasible actions $\mathcal A(s)$, and a text serialization of the current harness graph that lists each agent's role, tools, and outgoing edges. Because the feasible actions are part of the input, the policy head scores exactly the actions that can be taken. Inputs are limited to 2,048 tokens, and long queries retain their beginning and end. Both heads read the representation of the final token, $h_\theta(q,s)$.

\textbf{Batched tree search.}
Evaluating one leaf per architect call would leave the accelerator largely idle, so SHIFT evaluates up to eight leaves in a single forward pass. Without further care, parallel descents would all follow the same promising branch. We prevent this with a virtual loss: while a leaf awaits evaluation, every node on its path counts it as a visit of value zero, which temporarily lowers the branch's score and steers other descents elsewhere. Writing $L(s)$ and $L(s,a)$ for these pending visits and $W(s,a)$ for the sum of values backed up through action $a$, the statistics in Equation~(\ref{eq:puct}) are $\widetilde N(s)=N(s)+L(s)$, $\widetilde N(s,a)=N(s,a)+L(s,a)$, and
\begin{equation}
\widetilde Q(s,a)=
\begin{cases}
W(s,a)/\widetilde N(s,a), & \widetilde N(s,a)>0,\\
\operatorname{clip}\big(Q(s)-0.1,\,0,\,1\big), & \widetilde N(s,a)=0,\ N(s)>0,\\
0.5, & \text{otherwise},
\end{cases}
\end{equation}
where $Q(s)$ is the mean value backed up through $s$. Untried actions are thus initialized slightly below their parent's value, a conservative choice that keeps search on actions the policy favors until alternatives are evaluated. Virtual visits are removed as soon as the true values are backed up.

\textbf{Execution reward.}
The value head learns to predict a utility that rewards task success and penalizes the resources a harness consumes, so that search favors the least expensive harness that solves the query. For an executed harness with task score $u\in[0,1]$ (binary success when the benchmark provides no graded score), the reward and its rescaled utility are
\begin{equation}
 R=\alpha u-b-\sum_{x}\gamma_x r_x,
 \qquad z=\frac{\operatorname{clip}(R,-1,1)+1}{2},
 \label{eq:reward}
\end{equation}
where $z\in[0,1]$ is the value head's target and the sum runs over five penalties, each in $[0,1]$. With $t$ and $\ell$ the execution tokens and latency, and $\rho_{t}$ and $\rho_{\ell}$ benchmark-specific reference scales (Table~\ref{tab:reward_scales}),
\begin{align}
r_{\mathrm{tok}}&=\frac{t}{t+\rho_{t}}, &
r_{\mathrm{lat}}&=\frac{\ell}{\ell+\rho_{\ell}}, &
r_{\mathrm{tool}}&=\min\!\left(1,\tfrac{n_{\mathrm{tool}}}{6}\right), &
r_{\mathrm{agent}}&=\min\!\left(1,\tfrac{|\mathcal V|-1}{7}\right),
\end{align}
and $r_{\mathrm{timeout}}\in\{0,1\}$ indicates a timeout. The token and latency penalties saturate smoothly relative to their reference scales, which places benchmarks of very different cost, from GSM8K to OfficeQA, on a common scale while retaining sensitivity to cost beyond the reference. The tool penalty counts the distinct tool types granted, $n_{\mathrm{tool}}$, rather than tool calls, so it charges for the capabilities a harness is given; tools essential to a benchmark, such as the file reader for OfficeQA, are exempt. The agent penalty grows with each agent beyond the first, $|\mathcal V|$ being the number of agents. We use $\alpha=1.5$, $b=0.5$, and weights $\gamma$ of $0.15$ (tokens), $0.10$ (latency), $0.10$ (tools), $0.08$ (agents), and $0.15$ (timeout). Since the penalties sum to at most $0.58$, every successful harness receives a utility of at least $0.71$ and every failed harness at most $0.25$: success always dominates, and among harnesses that succeed, the cheaper one is preferred.

\begin{table}[t]
\centering\small
\caption{Training hyperparameters for the Gemma 4 E2B architect.}
\label{tab:train_settings}
\begin{tabular}{lr}
\toprule
Setting & Value\\
\midrule
LoRA rank / scale / dropout & 16 / 32 / 0.05\\
LoRA target layers & attention query, key, value, output\\
Optimizer & AdamW, learning rate $10^{-4}$, weight decay 0.01\\
Learning-rate schedule & cosine, minimum $10^{-5}$\\
Gradient clipping & 1.0\\
Epochs & 4\\
Entropy weight $\beta$ / value weight $\lambda_v$ / ranking weight $\lambda_r$ & 0.01 / 1 / 1\\
Ranking margin $m$ / minimum utility gap $\delta$ & 0.05 / 0.05\\
Root exploration noise & Dirichlet(0.35), mixing weight 0.25\\
Replay buffer & 256 queries, prioritized (exponents 0.6 / 0.4)\\
Minibatch size & 8\\
Other harnesses executed per query, $K_{\mathrm{exec}}$ & 4\\
\bottomrule
\end{tabular}
\end{table}

\begin{algorithm}[tbp]
\caption{Training the SHIFT architect}
\label{alg:train}
\begin{algorithmic}[1]
\Require training queries $\mathcal Q_d$, architect $\theta$, executor, seed harness $s_0$, budgets $B,H,K_{\mathrm{exec}}$
\State $\mathcal Z\gets\varnothing$ \Comment{executed harnesses and utilities}
\For{each epoch}
 \For{$q\in\mathcal Q_d$ in random order}
  \State $\mathcal T\gets$ search from $s_0$ with budget $B$ and root noise (Algorithm~\ref{alg:search})
  \State sample a path by visit counts; record policy targets along it
  \State execute the sampled harness and up to $K_{\mathrm{exec}}$ other distinct harnesses from $\mathcal T$; compute utilities $z$ (Equation~\ref{eq:reward})
  \State add the executed harnesses to $\mathcal Z$ and update $\theta$ with $\mathcal L_{\mathrm{online}}$
 \EndFor
 \For{each replay step}
  \State sample executed harnesses and same-query pairs from $\mathcal Z$
  \State update $\theta$ with the value loss plus $\lambda_r\mathcal L_{\mathrm{rank}}$
 \EndFor
\EndFor
\State \Return $\theta$
\end{algorithmic}
\end{algorithm}

\subsection{Training}
\label{app:training}

\begin{figure}[tbp]
\centering
\includegraphics[width=0.5\linewidth]{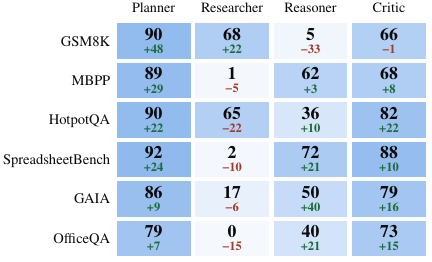}
\caption{How harness composition changes over training. Each cell gives the share (\%) of final-epoch training harnesses that contain a role (Verifier and Aggregator omitted), with the change since the first epoch below. Gemini 3.5 Flash executor.}
\label{fig:role_shift}
\end{figure}

Training alternates between search and learning (Algorithm~\ref{alg:train}). For each training query, the architect searches from $s_0$ with Dirichlet noise added to the root priors to encourage exploration, samples a construction path in proportion to visit counts, and executes the harness at its end together with up to $K_{\mathrm{exec}}$ other distinct harnesses recovered from the most-visited branches. As in AlphaZero~\cite{silver2017mastering}, the visit distributions along the path supervise the policy, while each executed harness supplies a measured utility for the value head. Executing several harnesses per query is what enables the ranking objective, since it yields pairs of harnesses for the same query whose utilities can be compared. Examples enter a prioritized replay buffer that favors states whose values are currently mispredicted, and the architect is updated online with $\mathcal L_{\mathrm{online}}$ (Equation~\ref{eq:online}). At the end of each epoch, a replay phase revisits all executed harnesses and optimizes the value loss jointly with $\mathcal L_{\mathrm{rank}}$ (Equation~\ref{eq:rank}) on same-query pairs. Table~\ref{tab:train_settings} lists the hyperparameters. Training also reshapes the harnesses the architect builds (Figure~\ref{fig:role_shift}): by the final epoch, Planners appear in most harnesses on every benchmark, while Researchers are largely dropped on MBPP, SpreadsheetBench, and OfficeQA and Reasoners become common on GAIA.

\textbf{Training cost.} Each architect trains for four epochs, and each epoch searches every training question once and executes up to five distinct harnesses from its tree. This amounts to 1,092 executions on OfficeQA, 1,611 on GAIA, 2,400 on MBPP, 4,250 on SpreadsheetBench, and about 9,400 each on GSM8K and HotpotQA. In executor tokens, training one architect costs about as much as answering 440 (OfficeQA) to 6,200 (HotpotQA) test questions with SHIFT-search. This cost is paid once per benchmark and executor and is then amortized over every later query, which each need only one execution.
\begin{figure}[tbp]
\centering
\includegraphics[width=0.78\linewidth]{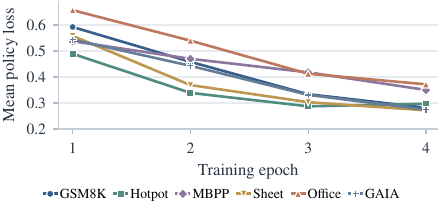}
\caption{Mean policy loss over four training epochs, one run per benchmark, averaging records with nonzero total loss. Hotpot, Sheet, and Office denote HotpotQA, SpreadsheetBench, and OfficeQA.}
\label{fig:training_dynamics}
\end{figure}

\textbf{Training curves.} Figure~\ref{fig:training_dynamics} tracks the policy objective over training. Each architect sees every training question once per epoch, with the same questions in all four epochs (Table~\ref{tab:benchmarks}), and we report the mean policy loss over the updates of each epoch. The loss falls by 35\% to 52\% on every benchmark. Because the policy targets are produced by search with the current architect, they sharpen as the architect improves; the decreasing loss therefore shows the policy increasingly anticipating the outcome of its own search, which is what allows search to concentrate its budget on promising actions.

\section{Harness Selection Strategies}
\label{app:decoding}

The architect provides two learned signals, a policy over actions and a value over harnesses, and a harness can be decoded from them in several ways, much as a language model can be decoded greedily or with search. We compare five decoding strategies that share the same trained architect for each benchmark, start from $s_0$, and apply at most ten actions. They differ only in which signal chooses each action and whether actions are chosen with lookahead.

\begin{itemize}[nosep]
\item \textbf{Greedy} applies the policy's most probable action at each step until the policy chooses to stop. It isolates the policy.
\item \textbf{Policy-veto} lets the policy propose actions in order of probability and the value head accept the first of the top four whose predicted utility is at least 0.55, falling back to the policy's first choice. The value can thus overrule the policy, but only one step at a time.
\item \textbf{Value-greedy} scores every one-action change with the value head and moves to the best one while it improves the current harness. It isolates the value without lookahead.
\item \textbf{SHIFT-search} runs Monte Carlo tree search with 64 simulations and executes the harness at the end of the most-visited path, combining both signals with lookahead.
\item \textbf{SHIFT-value} runs the same search, recovers up to five distinct harnesses from the most-visited branches, and executes the one with the highest predicted utility. If that utility is below 0.55, it searches once more with 192 additional simulations and chooses among both sets of candidates.
\end{itemize}

Comparing these strategies separates the contributions of the policy, the value, and lookahead (Table~\ref{tab:shift_modes}). Search matches the accuracy of greedy construction while using 30\% fewer execution tokens, because it weighs each action by the utility of the harnesses it leads to. Value-greedy is the cheapest strategy but loses about 20\% in mean accuracy: an agent or tool whose benefit appears only after later actions looks unattractive one step ahead. The value head is therefore most effective when combined with lookahead, as in SHIFT-search, or used to rank complete harnesses, as in SHIFT-value. Figure~\ref{fig:role_tokens} shows where search saves tokens relative to greedy construction.

\subsection{Cost savings across questions}
\label{app:paired_costs}

\begin{figure}[tbp]
\centering
\includegraphics[width=0.92\linewidth]{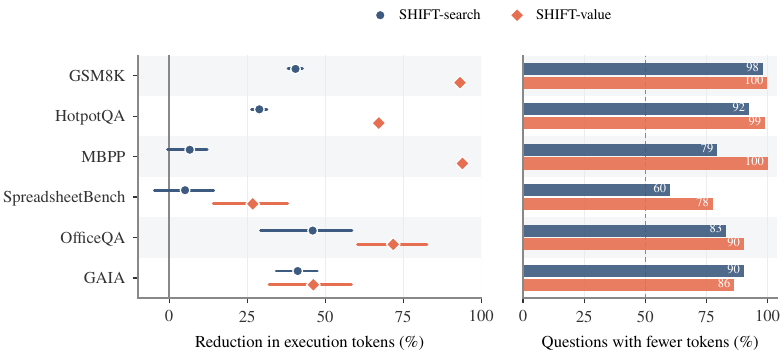}
\caption{Execution-token savings relative to greedy construction, paired by question. Left: reduction in mean execution tokens, with 95\% bootstrap intervals over questions. Right: share of questions on which each strategy uses fewer tokens than greedy construction.}
\label{fig:paired_costs}
\end{figure}

Aggregate savings could in principle come from a few very expensive questions. To test whether they are broad, we pair every evaluation question across strategies and compare the mean execution tokens of its three runs (Figure~\ref{fig:paired_costs}). Search uses fewer tokens than greedy construction on 60\% to 98\% of questions, depending on the benchmark, and SHIFT-value does so on 77.5\% to 100\%. The savings are largest on the most expensive benchmarks, where search cuts mean tokens by 46\% on OfficeQA and 41\% on GAIA.

\begin{figure}[tbp]
\centering
\includegraphics[width=\linewidth]{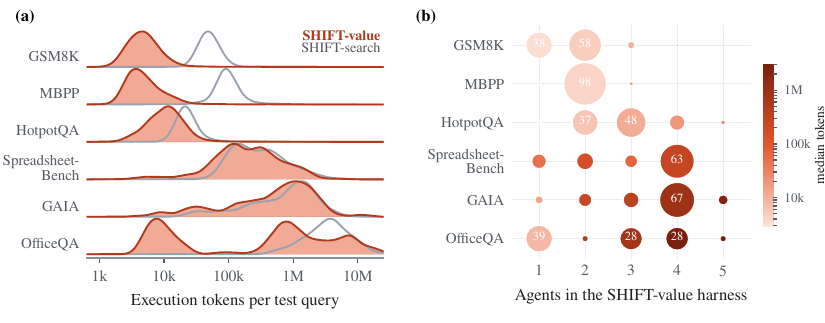}
\caption{Cost and harness size per test query (all test runs). (a) Distribution of execution tokens per query for SHIFT-search and SHIFT-value. (b) Harnesses chosen by SHIFT-value: bubble area is the share of test queries whose harness has that many agents (labels for shares of at least 25\%), and color is the median execution tokens of those queries.}
\label{fig:cost_size}
\end{figure}

\textbf{Cost and harness size per query.} Figure~\ref{fig:cost_size} shows how the two operating points spend tokens query by query. SHIFT-value spends markedly less than SHIFT-search on the short-horizon benchmarks and chooses among harnesses of different sizes for each query: on GSM8K and MBPP it answers almost every query with one or two agents at about a tenth of SHIFT-search's cost, on HotpotQA it mostly uses two or three, and on SpreadsheetBench and GAIA it builds harnesses with four or more agents, as search does. On OfficeQA its choices split between a single agent and three or four agents, which gives its cost distribution two modes.

\subsection{Architect overhead}
\label{app:inference_latency}

\begin{figure}[t]
\centering
\includegraphics[width=0.92\linewidth]{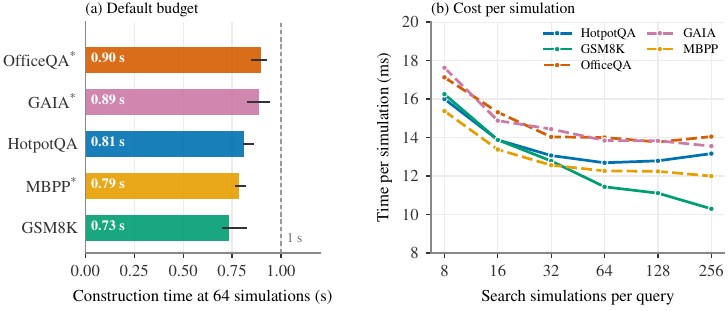}
\caption{Architect construction cost on one H100 NVL GPU (median over ten questions, three repeats each). (a) Time to construct one harness at the default budget of 64 simulations; whiskers show the interquartile range. (b) Time per simulation as the budget grows. Benchmarks marked $^{*}$ (dashed in b) use the HotpotQA architect with the benchmark's tools.}
\label{fig:architect_latency}
\end{figure}

Because SHIFT replaces executor calls with local inference, its design cost per query is the time the architect spends searching. We measure the complete construction call on a single H100 NVL GPU for ten questions on each of five benchmarks, with three repeats each and model loading excluded (Figure~\ref{fig:architect_latency}). At the default budget of 64 simulations, constructing a harness takes a median of 0.81 seconds on HotpotQA and 0.73 seconds on GSM8K, and between 0.79 and 0.90 seconds on OfficeQA, GAIA, and MBPP. Greedy construction takes 0.26 seconds, and SHIFT-value, including its occasional second search, 0.86 and 0.78 seconds on HotpotQA and GSM8K. Larger budgets remain affordable: each simulation costs 10 to 14 milliseconds at budgets of 64 and above, slightly less than at small budgets as fixed per-call overhead is amortized, so construction time grows linearly with the budget and reaches 2.6 to 3.6 seconds at 256 simulations. Because simulations are evaluated in batches by the Gemma 4 E2B model, the design cost stays below one second per query at the default budget, whereas executing a single multi-agent harness on OfficeQA or GAIA consumes roughly one to eight million executor tokens (Figure~\ref{fig:role_tokens}).

\section{Benchmarks and Evaluation Protocol}
\label{app:reporting}

\textbf{Benchmark selection.} The six benchmarks are chosen to span the range over which the value of a harness should vary. At one end, GSM8K, HotpotQA, and MBPP pose short, self-contained problems that a single well-prompted agent largely solves. At the other, SpreadsheetBench, OfficeQA, and GAIA require reading and producing files, querying large document collections, or browsing the web over many steps, so success depends on which agents are present and which tools they can use. Evaluating on both ends tests whether SHIFT adds structure where it helps and withholds it where it does not. Table~\ref{tab:benchmarks} summarizes each benchmark, its success criterion, and its data split.

\begin{table}[tbp]
\centering\small
\caption{Benchmarks, success criteria, and data splits (number of tasks). Training, validation, and test sets are disjoint.}
\label{tab:benchmarks}
\setlength{\tabcolsep}{4pt}
\begin{tabularx}{\textwidth}{@{}l>{\raggedright\arraybackslash}X>{\raggedright\arraybackslash}Xrrr@{}}
\toprule
Benchmark & Task & Success criterion & Train & Val & Test\\
\midrule
GSM8K & Grade-school math word problems & Final number equals the reference & 472 & 128 & 1,319\\
HotpotQA & Multi-hop questions over supplied passages (distractor setting) & Exact match after standard answer normalization & 472 & 128 & 7,405\\
MBPP & Short Python functions from a description (sanitized set) & Generated code passes all unit tests & 120 & 43 & 257\\
SpreadsheetBench & Spreadsheet manipulation from natural-language instructions & Output workbook matches the reference on every test case & 220 & 60 & 120\\
OfficeQA Pro & Questions over a corpus of 697 U.S.\ Treasury Bulletins & Answer matches the reference within 1\% & 73 & 19 & 41\\
GAIA & General assistant tasks needing web search, files, and reasoning & Normalized answer matches the reference & 90 & 24 & 51\\
\bottomrule
\end{tabularx}
\end{table}

GSM8K uses its official test set (1,319 questions) and HotpotQA its official distractor development set (7,405), since HotpotQA test labels are not public; from each official training set we draw 600 questions, 472 for training and 128 for validation. MBPP keeps its official splits. SpreadsheetBench (400 tasks), OfficeQA Pro (133 questions), and GAIA (the 165 tasks of its public validation release) provide no training split, so we divide each at random into training, validation, and test sets in proportions of 55\%, 15\%, and 30\%. Test questions are never used for training or model selection, and all methods are evaluated on the same test sets.

\textbf{Metrics.} Accuracy is each benchmark's success criterion averaged over the three test runs. Every test task counts: failures and timeouts are scored as incorrect and are never removed. The mean in Table~\ref{tab:main} weights benchmarks equally, so that HotpotQA's 7,405 questions do not outweigh OfficeQA's 41. Execution cost counts the tokens consumed by the executor across all agents of the executed harness, relative to Direct. Because this ratio spans about two orders of magnitude across methods, we aggregate it with a geometric mean, which weights a halving of cost equally on every benchmark. Architect computation is reported separately (Appendix~\ref{app:inference_latency}).

\section{Baselines}
\label{app:baseline_provenance}

The baselines fall into the four groups of Table~\ref{tab:main}. Prompting methods fix a single-agent harness. Tool-using and multi-agent frameworks (ReAct, LAP, and AgentVerse) fix a harness with tools or several agents; LAP (LLM as a Policy) runs a tree search over tool-using actions in which the executor LLM proposes and scores the actions. The DSPy programs run unoptimized or with instructions and demonstrations tuned by MIPROv2 or GEPA. Workflow optimizers (AFlow, GPTSwarm, and Trace) search for one structure per benchmark, and DyLAN composes its agent team for each query. SHIFT differs from all of them in constructing the structure, instructions, and tool access of the harness jointly for each query.

To isolate the effect of harness design, every baseline runs in the same execution and grading framework as SHIFT, with the same Gemini 3.5 Flash executor, the same test sets, and the same answer checking. Differences in Table~\ref{tab:main} therefore reflect how each method organizes the executor rather than the model or the grader. Every cell averages three runs, except GPTSwarm on HotpotQA and OfficeQA, which average two.

\textbf{Zero scores on SpreadsheetBench.} A SpreadsheetBench task counts as solved only if code writes a modified workbook to disk and that workbook matches the reference on every test case. The task therefore cannot be completed without file and code-execution tools. Methods that return their answer as text (Direct, CoT (few-shot), Self-Refine, Tree-of-Thoughts, and our AFlow configuration) never produce the output workbook and score zero. CoT, Self-Consistency, and Plan-and-Solve also answer in text, so we assign them zero without running them; the zero enters their Mean, and their Cost excludes SpreadsheetBench ($^{\ddagger}$ in Table~\ref{tab:main}). Methods that can equip agents with file and code tools, including ReAct, the DSPy programs, and SHIFT, can solve these tasks.

\section{Additional Results on Harness Accuracy}
\label{app:rq1}

This appendix supports the accuracy results of Section~\ref{sec:main}. We first show that the benefit of structure differs across benchmarks, which is the premise of per-query construction; we then test the gains over the strongest baseline question by question, and finally examine which parts of the action space the gains require.

\subsection{Structure sensitivity}
\label{app:structure_gain}

Per-query construction is only worthwhile if the right amount of structure differs from task to task. We test this premise on the training runs. During training, search executes many different harnesses on the same questions, which yields a large and varied sample of harnesses run by the Gemini 3.5 Flash executor. For each benchmark, we group these executions by their amount of structure along one action type and compare success between the two extremes: one agent versus four or more, no tools versus every tool the benchmark offers, and no appended directives versus two or more. Figure~\ref{fig:structure_gain} reports the resulting change in utility, and Table~\ref{tab:structure_gain} gives the success rates and group sizes. All nine contrasts on SpreadsheetBench, GAIA, and OfficeQA are significant (two-sided Fisher exact test, $p\le0.016$), and none on GSM8K. The effect of structure ranges from none on GSM8K to 50\% on OfficeQA, and it differs by action type within a benchmark: on SpreadsheetBench, tools matter most, and on OfficeQA, agents. Because search chose which harnesses to run, this comparison is observational: it shows where structure and success go together rather than the effect of adding structure to a fixed harness.

\begin{table}[tbp]
\centering\small
\caption{Success rates (\%) of executions with the least and most of each structure type; the number of executions is in parentheses. $^{*}$The two groups differ at $p<0.05$ (two-sided Fisher exact test).}
\label{tab:structure_gain}
\setlength{\tabcolsep}{4pt}
\begin{tabular}{lrrrrrr}
\toprule
 & \multicolumn{2}{c}{Agents} & \multicolumn{2}{c}{Tools} & \multicolumn{2}{c}{Directives}\\
\cmidrule(lr){2-3}\cmidrule(lr){4-5}\cmidrule(l){6-7}
Benchmark & 1 & $\ge4$ & none & all & 0 & $\ge2$\\
\midrule
GSM8K & 98.2 (114) & 98.4 (835) & 98.4 (127) & 97.8 (965) & 98.0 (665) & 97.9 (419)\\
MBPP & 94.1 (34) & 99.0 (210) & 95.6 (68) & 99.4$^{*}$ (319) & 98.0 (247) & 100.0 (37)\\
HotpotQA & 75.6 (90) & 84.3$^{*}$ (1,116) & 72.4 (76) & 83.7$^{*}$ (1,611) & 79.6 (554) & 83.1 (278)\\
SpreadsheetBench & 55.3 (38) & 82.8$^{*}$ (476) & 44.8 (29) & 87.3$^{*}$ (700) & 65.4 (208) & 83.0$^{*}$ (406)\\
GAIA & 40.0 (25) & 73.9$^{*}$ (134) & 35.3 (17) & 66.9$^{*}$ (290) & 50.0 (90) & 72.2$^{*}$ (158)\\
OfficeQA & 9.4 (32) & 59.3$^{*}$ (59) & 16.0 (25) & 46.8$^{*}$ (190) & 26.2 (61) & 54.6$^{*}$ (97)\\
\bottomrule
\end{tabular}
\end{table}

\subsection{Paired comparison with Trace}
\label{app:paired_main}

A higher mean accuracy can arise from a few benchmarks or a few questions, so we compare SHIFT-search with Trace, the strongest baseline, question by question. For each benchmark, we average the three runs of each question and resample questions with replacement (20,000 bootstrap draws), weighting benchmarks equally in the six-benchmark mean. SHIFT-search improves the mean by 7.2\% (95\% interval [4.4, 9.9]), well above zero. The improvement comes from OfficeQA (+30.1\%, [17.1, 43.1]) and GAIA (+15.0\%, [5.9, 24.8]), while the two methods are within about one percent on the other four benchmarks.
\begin{figure}[tbp]
\centering
\includegraphics[width=0.95\linewidth]{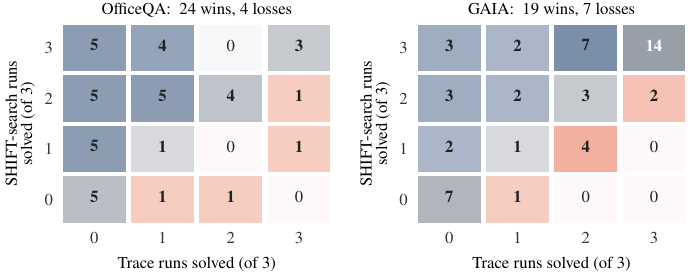}
\caption{Question-level comparison with Trace, the strongest baseline, on OfficeQA (41 questions) and GAIA (51). Each cell counts the questions that SHIFT-search (rows) and Trace (columns) solve in the given number of their three runs; cells above the diagonal are questions SHIFT-search solves more often.}
\label{fig:paired_trace}
\end{figure}

The gains are also broad within these benchmarks (Figure~\ref{fig:paired_trace}). Counting successes over three runs, SHIFT-search solves more often than Trace on 24 OfficeQA questions and less often on only 4, and on GAIA the counts are 19 and 7; SHIFT-value likewise wins more questions than it loses on both benchmarks (15 against 9 on OfficeQA and 18 against 7 on GAIA). SHIFT-search also solves five OfficeQA questions and three GAIA questions in every run on which Trace fails in every run, with no case in the opposite direction.

\textbf{Robustness of the ranking.}
Two alternative aggregations leave the ranking unchanged. Excluding SpreadsheetBench, which three baselines did not run, SHIFT-search reaches 78.3\% and SHIFT-value 73.0\%, against 69.6\% for Trace, the strongest baseline. Removing OfficeQA, the benchmark with the largest gain, SHIFT-search still leads with 84.0\%.

\subsection{Action-space restrictions}
\label{app:edit_spaces}

\begin{table}[tbp]
\centering\small
\caption{HotpotQA accuracy under restricted action spaces (500 questions, mean $\pm$ SD over three runs; change with paired 95\% bootstrap interval).}
\label{tab:edit_spaces}
\begin{tabular}{lrr}
\toprule
Actions allowed & Accuracy (\%) & Change vs.\ full (\%)\\
\midrule
All actions & $69.4_{\pm0.5}$ & --\\
Tool grants only & $67.9_{\pm0.3}$ & $-1.5$ [$-3.1$, $-0.1$]\\
Directives only & $60.3_{\pm0.9}$ & $-9.1$ [$-12.0$, $-6.4$]\\
\bottomrule
\end{tabular}
\end{table}

To test whether the gains require choosing structure, instructions, and tools together, we restrict the actions available to the trained HotpotQA architect at inference time, without retraining, on 500 HotpotQA test questions with three runs each and 64 simulations. \emph{Directives only} allows only instruction actions (actions 10--25 in Table~\ref{tab:action_catalogue}), and \emph{tool grants only} allows only tool actions (actions 10 and 26--40); both keep the single Coder, since they cannot add agents. Each restriction lowers accuracy (Table~\ref{tab:edit_spaces}). Paired over questions with 20,000 bootstrap draws, directives alone lose 9.1\% (95\% interval [6.4, 12.0]) and tool grants alone lose 1.5\% ([0.1, 3.1]); both intervals exclude zero. Tools recover most of the accuracy on this retrieval-heavy benchmark, but only the full action space, which can also add agents and instructions, reaches the best result.

\section{Learned Value for Harness Selection}
\label{app:rq3}

This appendix supports Section~\ref{sec:controls}. SHIFT-value relies on the value head to choose among candidate harnesses without executing them, so we evaluate the value head directly as a ranker, separately from the search that proposes candidates.

\subsection{Candidate ranking}
\label{app:ranking_protocol}

\textbf{Setup.} Evaluating a ranker requires knowing how every candidate would have performed. We therefore build a pool of distinct candidate harnesses for each of 400 HotpotQA and 200 GSM8K questions from several construction strategies (greedy construction, search with 8 and 64 simulations, search with a constant value, and a fixed reference harness), execute every candidate twice with a Gemma 3 27B executor, and score each candidate with the value head before any outcome is inspected. Each pool holds four or five candidates. For question $i$ with pool $\mathcal C_i$, let $v_{ic}$ be the predicted utility of candidate $c$ and $y_{ic}$ its mean success. Value selection and uniform selection achieve
\begin{align}
A_i^{\mathrm{value}}&=\frac{1}{|M_i|}\sum_{c\in M_i}y_{ic},\quad M_i=\operatorname*{arg\,max}_{c\in\mathcal C_i}v_{ic},
&
A_i^{\mathrm{uniform}}&=\frac{1}{|\mathcal C_i|}\sum_{c\in\mathcal C_i}y_{ic},
\end{align}
where ties in predicted utility are averaged. Uniform selection is computed in expectation, so it adds no sampling noise, and the two selectors are compared on identical pools and outcomes.

\begin{figure}[tbp]
\centering
\includegraphics[width=0.95\linewidth]{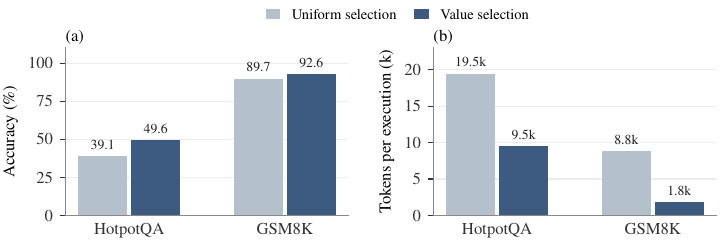}
\caption{Choosing among the same candidate harnesses with the value head versus uniformly at random (400 HotpotQA and 200 GSM8K questions, Gemma 3 27B executor). (a) Accuracy. (b) Execution tokens of the selected harness.}
\label{fig:value_selection}
\end{figure}

\textbf{The value head selects more accurate and cheaper harnesses.} Value selection outperforms uniform selection by 10.5\% on HotpotQA (95\% bootstrap interval over questions [7.5, 13.6]) and by 2.9\% on GSM8K ([0.8, 5.0]). It does so while choosing much cheaper harnesses (Figure~\ref{fig:value_selection}): the selected harness uses 9.5k execution tokens on HotpotQA against 19.5k for a uniform choice, and 1.8k against 8.8k on GSM8K. Because the value head is trained on a utility that rewards success and charges for cost (Equation~\ref{eq:reward}), it learns to prefer the least expensive harness that still solves the query, which is the behavior that gives SHIFT-value its low execution cost.

\textbf{More candidates, better choices.} For Figure~\ref{fig:offline_ranking}, we average each selector over every subset of $k$ candidates from each pool, for $k=1,\dots,4$, so that every question contributes at every $k$. Uniform selection has the same expected accuracy for every $k$, so a rising value curve reflects better choices rather than easier questions. Value selection improves steadily as more candidates become available, from 39.1\% with one candidate to 42.7\%, 46.1\%, and 48.9\% with four on HotpotQA, and from 89.7\% to 92.4\% on GSM8K.

\section{Transfer to Harder Tasks}
\label{app:rq4}

This appendix supports Section~\ref{sec:executor_transfer}. A useful architect should carry over to tasks harder than those it was trained on; we test this by applying the GSM8K architect, without retraining, to MATH-500.

\label{app:math_transfer}

We apply the GSM8K architect, without further training and without access to MATH answers, to all 500 problems of MATH-500~\cite{hendrycks2021math,lightman2024verify}, which span seven subjects and five difficulty levels. No MATH-500 question appears in the GSM8K splits. Each problem is executed once with Llama 4 Scout at temperature 0.2, and answers are graded with the Math-Verify library~\cite{kydlicekmathverify}.

\begin{figure}[tbp]
\centering
\includegraphics[width=0.5\linewidth]{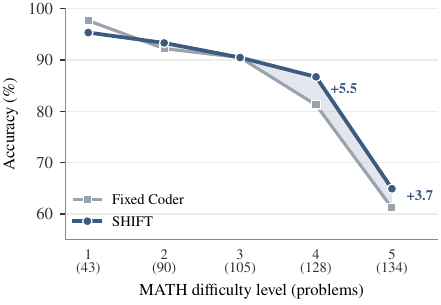}
\caption{MATH-500 accuracy by difficulty level for SHIFT, using the GSM8K architect without retraining.}
\label{fig:transfer}
\end{figure}

SHIFT reaches 83.6\%, against 81.2\% for a fixed single Coder, and the gain is concentrated where it should be (Figure~\ref{fig:transfer}). On difficulty levels 1--3, which a single agent already solves about 92\% of the time, the two are tied. On the two hardest levels, SHIFT improves accuracy by 5.5\% and 3.7\%, reaching 75.6\% against 71.0\% over these 262 problems; paired over problems, this difference of 4.6\% has a 95\% bootstrap interval of [0.4, 9.2]. The harnesses built by an architect trained only on grade-school arithmetic thus help most on the hardest problems, where a single agent falls short. Table~\ref{tab:math_transfer} further shows that the complete architect, with both its learned policy and its learned value, is the most accurate configuration: replacing either with an uninformed version lowers accuracy.

\begin{table}[htbp]
\centering\small
\caption{MATH-500 accuracy with Llama 4 Scout, using the GSM8K architect without retraining (500 problems per row, one execution each). Uniform policy: feasible actions receive equal prior probability. Constant value: every harness receives a predicted utility of 0.5.}
\label{tab:math_transfer}
\begin{tabular}{lrr}
\toprule
Configuration & Accuracy (\%) & Executor tokens / problem\\
\midrule
SHIFT (learned policy, learned value) & \textbf{83.6} & 8,049\\
Learned policy, constant value & 82.4 & 7,614\\
Uniform policy, constant value & 82.0 & 6,283\\
Uniform policy, learned value & 81.2 & 5,585\\
Fixed single Coder & 81.2 & 1,064\\
\bottomrule
\end{tabular}
\end{table}